\documentclass[sigconf]{acmart}

\usepackage{multirow}
\usepackage{colortbl}
\usepackage{amsmath}

\usepackage{amssymb}
\usepackage{bm}

\AtBeginDocument{%
  }

\acmConference[ASP-DAC'27]{32nd Asia and South Pacific Design Automation Conference}{Jan. 25-28, 2027}{Tokyo, Japan}

\begin{document}

\title{ASSERT: Adaptive Stochastic Sampling for Robust Diffusion Models on Analog Compute-in-Memory Hardware}

\author{Yuannuo Feng}
% \authornote{Equal contribution.}
\affiliation{%
  \institution{School of Integrated Circuit Science and Engineering, Beihang University}
  \city{Beijing}
  \country{China}
}
\affiliation{
 \institution{Zhicun Research Lab}
 \city{Beijing}
 \country{China}
}

\author{Yizhe Chen}
\affiliation{%
  \institution{School of Integrated Circuit Science and Engineering, Beihang University}
  \city{Beijing}
  \country{China}
}
\affiliation{
 \institution{Zhicun Research Lab}
 \city{Beijing}
 \country{China}
}

\author{Wenshuai Yao}
% \authornotemark[1]
\affiliation{
\institution{School of Integrated Circuits, Peking University}
\city{Beijing}
\country{China}
}
\affiliation{
 \institution{Zhicun Research Lab}
 \city{Beijing}
 \country{China}
}

\author{Yuxin Xie}
\affiliation{%
  \institution{School of Integrated Circuit Science and Engineering, Beihang University}
  \city{Beijing}
  \country{China}
}

\author{Ngai Wong}
\affiliation{%
  \institution{Department of Electrical and Computer Engineering, The University of Hong Kong}
  \city{Hong Kong}
  \country{China}
}

\author{Wenyong Zhou}
\authornote{Corresponding authors.}
\affiliation{%
  \institution{Department of Electrical and Computer Engineering, The University of Hong Kong}
  \city{Hong Kong}
  \country{China}
}
\affiliation{
 \institution{Zhicun Research Lab}
 \city{Beijing}
 \country{China}
}

\author{Wang Kang}
\authornotemark[1]
\affiliation{%
  \institution{School of Integrated Circuit Science and Engineering, Beihang University}
  \city{Beijing}
  \country{China}
}

\begin{abstract}
Diffusion models achieve strong image generation quality but incur high iterative denoising costs. Analog compute-in-memory (CIM) can accelerate matrix-vector multiplications, yet spatial memory variations perturb weights and accumulate during sampling. Unlike conventional neural networks, diffusion models' temporal sensitivity to hardware noise remains underexplored.
We investigate diffusion inference using a noise model calibrated and validated against measurements collected from multiple physical CIM chips. Our results show that the early, high-noise denoising stage is substantially more vulnerable than the final refinement stage. A first-order trajectory analysis attributes this behavior to the repeated propagation of correlated prediction errors induced by a fixed hardware mapping. Based on this observation, we propose ASSERT, a training-free sampler that uses higher stochasticity early and smoothly transitions to deterministic denoising. The injected stochasticity changes subsequent activation trajectories and thereby reduces their alignment with persistent spatial errors.
Across the evaluated settings, ASSERT achieves up to 2.58$\times$ lower FID than deterministic DDIM on high-resolution datasets and 7.68$\times$ lower FID in the CIFAR-10 step-count study, without changing model parameters or the number of network evaluations.
\end{abstract}

\begin{CCSXML}
<ccs2012>
   <concept>
       <concept_id>10010147.10010178.10010224</concept_id>
       <concept_desc>Computing methodologies~Computer vision</concept_desc>
       <concept_significance>500</concept_significance>
       </concept>
 </ccs2012>
\end{CCSXML}

\ccsdesc[500]{Computing methodologies~Computer vision}
\keywords{Diffusion models, Compute-in-memory, Hardware noise, Noise-aware sampling}
\maketitle

\section{Introduction}
\label{sec:introduction}
Diffusion models achieve strong performance across visual synthesis tasks through iterative denoising~\cite{DDPM,DDIM,GAN,vae,razavi2019generating}. However, sequential sampling can require hundreds of neural-network evaluations, whose matrix-vector multiplications (MVMs) dominate the computational workload~\cite{yang2023diffusion,lu2022dpm,qdiffusion}. This cost hinders efficient deployment as models grow.

To address these computational challenges, compute-in-memory (CIM) architectures have emerged as a compelling acceleration alternative for diffusion models. Recent advances demonstrate the compatibility between CIM hardware and diffusion models, delivering substantial improvements in energy efficiency and throughput by performing matrix-vector multiplications directly within memory arrays, thereby eliminating costly data movement between processing units and memory~\cite{pan2022mini,verma2019memory}. 

However, the inherent hardware imperfections in CIM architectures perturb stored weights, posing severe challenges for model performance~\cite{CIM_noise,wang202340nm,asicon,icassp}. These imperfections stem from various sources including device variability, programming uncertainty, thermal fluctuations, and quantization noise, which collectively introduce weight perturbations that can significantly degrade generation quality.

\begin{figure}[!t]
\centering
\includegraphics[width=1.0\columnwidth]{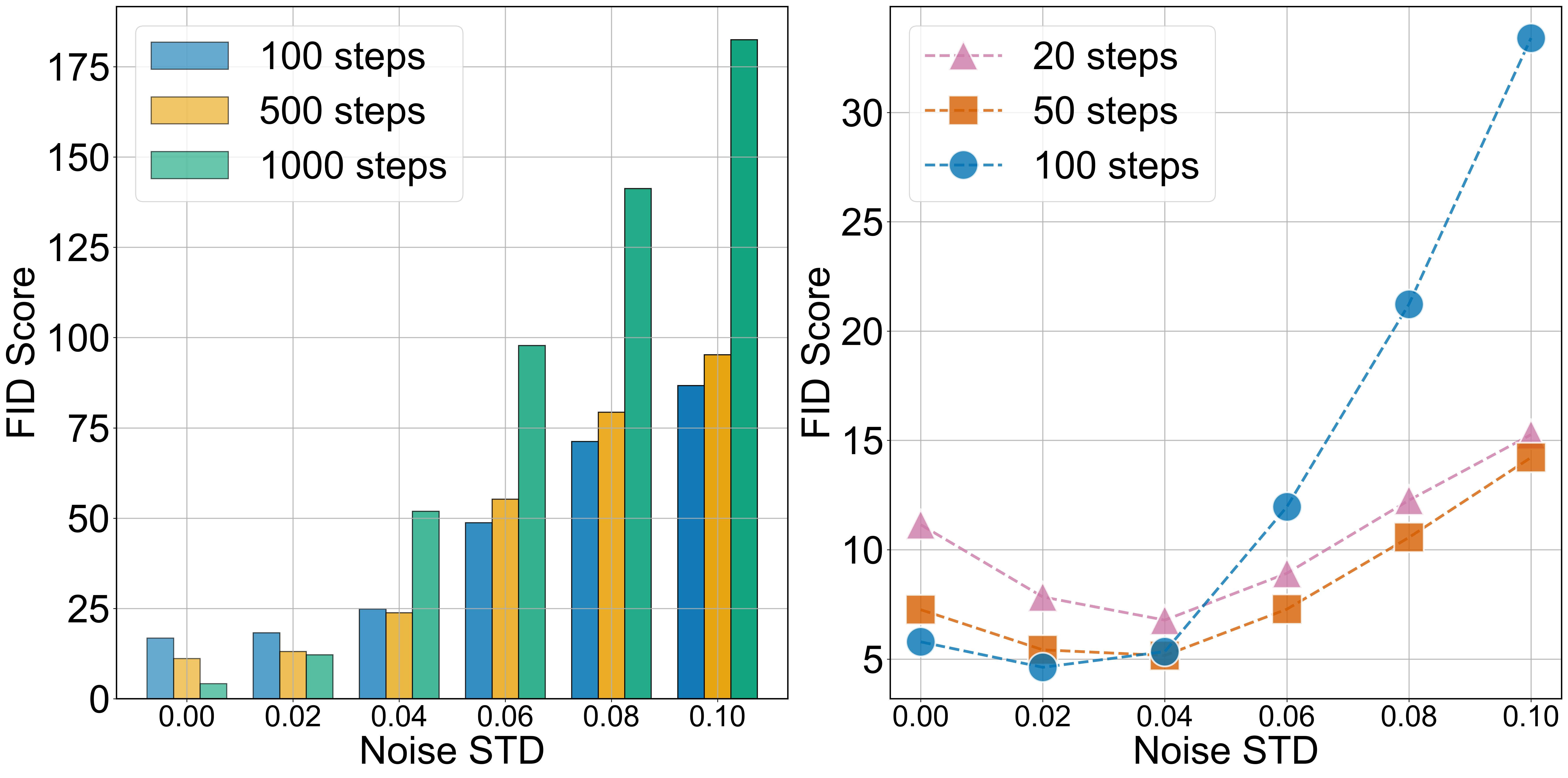}
\caption{CIFAR-10 FID under hardware noise for DDPM (left) and DDIM (right).}
% \vspace{-0.2cm}
\label{fig:motivation}
\end{figure}

\begin{figure}[!t]
\centering
\includegraphics[width=0.9\columnwidth]{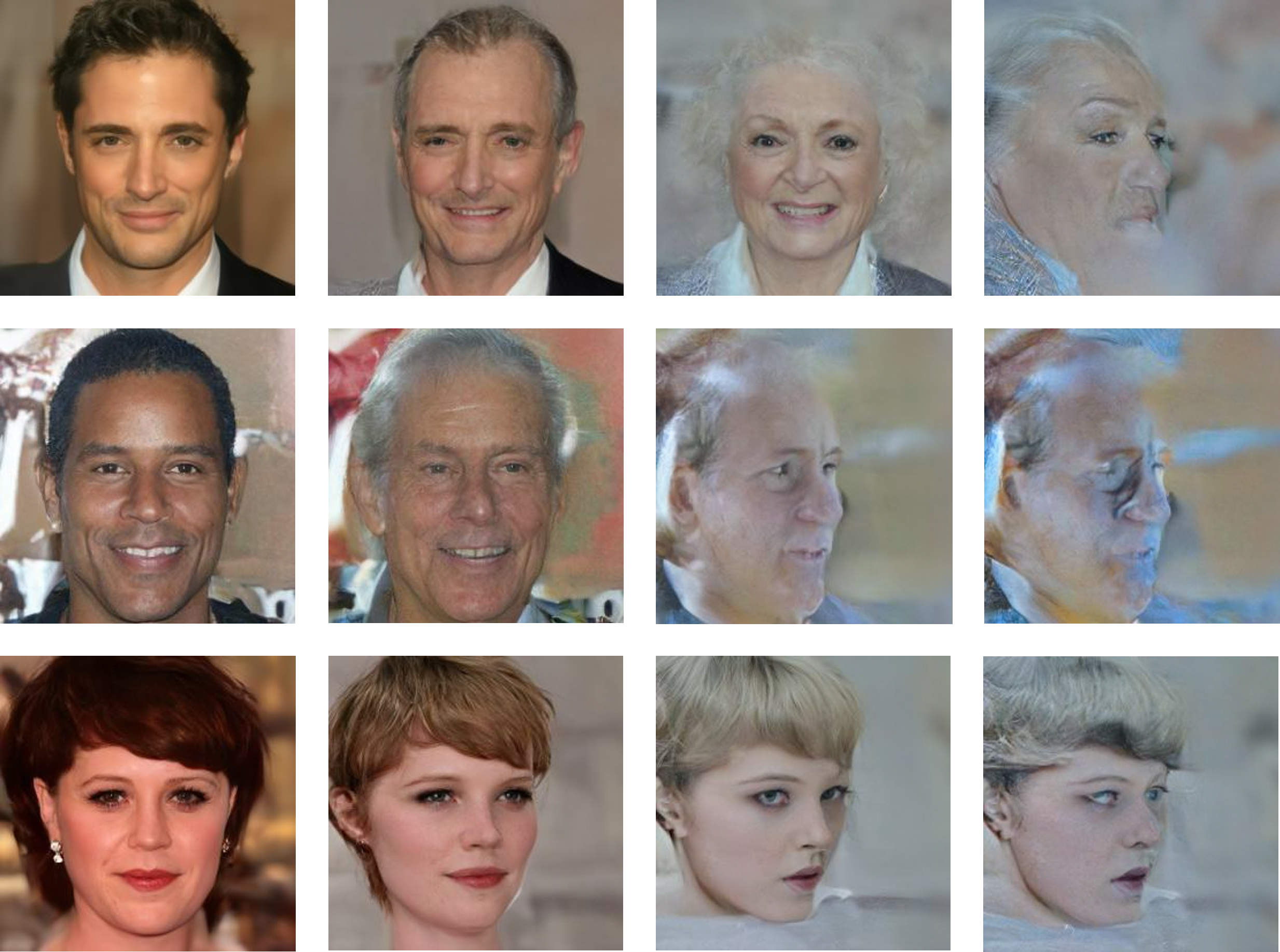}
\caption{CelebA-HQ samples from 100-step DDIM under increasing hardware noise.}
% \vspace{-0.2cm}
\label{fig:cele_noise}
\end{figure}

As shown in Figure~\ref{fig:motivation}, FID increases rapidly as weight perturbations intensify, revealing the vulnerability of iterative diffusion inference to hardware noise. Figure~\ref{fig:cele_noise} provides a qualitative view: facial structure progressively degrades into blurred texture and incoherent artifacts as noise increases.

Previous works have examined the robustness of neural networks under CIM-induced perturbations~\cite{asicon,mao2025hyimc}, yet existing solutions rely heavily on retraining or fine-tuning procedures that require substantial computational and hardware resources. Moreover, prior research primarily targets classification or language generation tasks and does not address the unique iterative characteristics of diffusion sampling. The behavior of diffusion models under analog hardware noise remains largely unexplored, limiting their practical integration into CIM-based systems.

%%%%%%%%%%%%%%%%%%%%%%%%
%%%%%%%%%%%%%%%%%%%%%%%%
%%%%%%%%%%%%%%%%%%%%%%%%
%%%%%%%%%%%%%%%%%%%%%%%%

To address this gap, we conduct an end-to-end diffusion-inference study using a noise model calibrated and validated with measurements from multiple physical CIM chips. Our analysis identifies temporal sensitivity patterns and quantifies the benefit of stochastic sampling. Building on these findings, we introduce ASSERT, a training-free sampling method that adaptively interpolates between stochastic and deterministic DDIM updates. The contributions of this work are summarized as follows:
\begin{itemize}
    \item We characterize diffusion-model robustness with a multi-chip-calibrated noise model and show that the early, high-noise denoising stage is more vulnerable than the final refinement stage.
    \item We derive a first-order error recursion for persistent spatial weight noise and propose ASSERT, which reduces cross-step error alignment by combining early stochastic exploration with deterministic refinement.
    \item Experiments across multiple datasets show up to 2.58$\times$ and 7.68$\times$ lower FID in the high-resolution and CIFAR-10 step-count evaluations, respectively.
\end{itemize}

%%%%%%%%%%%%%%%%%%%%%%%%%%%%%%%%%%%%%%%%%%%%%%
\section{Preliminary}
\label{sec:related}

\subsection{Diffusion Models }
Diffusion models learn to reverse a gradual noising process to synthesize data. DDPM~\cite{DDPM} formulates generation as the learned reverse of a fixed Markov chain, while DDIM~\cite{DDIM} exposes a non-Markovian family whose stochasticity parameter interpolates between deterministic and stochastic trajectories. Fast solvers and distillation further reduce network evaluations~\cite{PNDM,lu2022dpm,salimans2022progressive}. ASSERT does not introduce a new diffusion family; its novelty is a hardware-noise-aware temporal schedule for the existing DDIM stochasticity parameter, motivated by the propagation of persistent CIM errors.

\begin{figure}[!t]
\centering
\includegraphics[width=1.0\columnwidth]{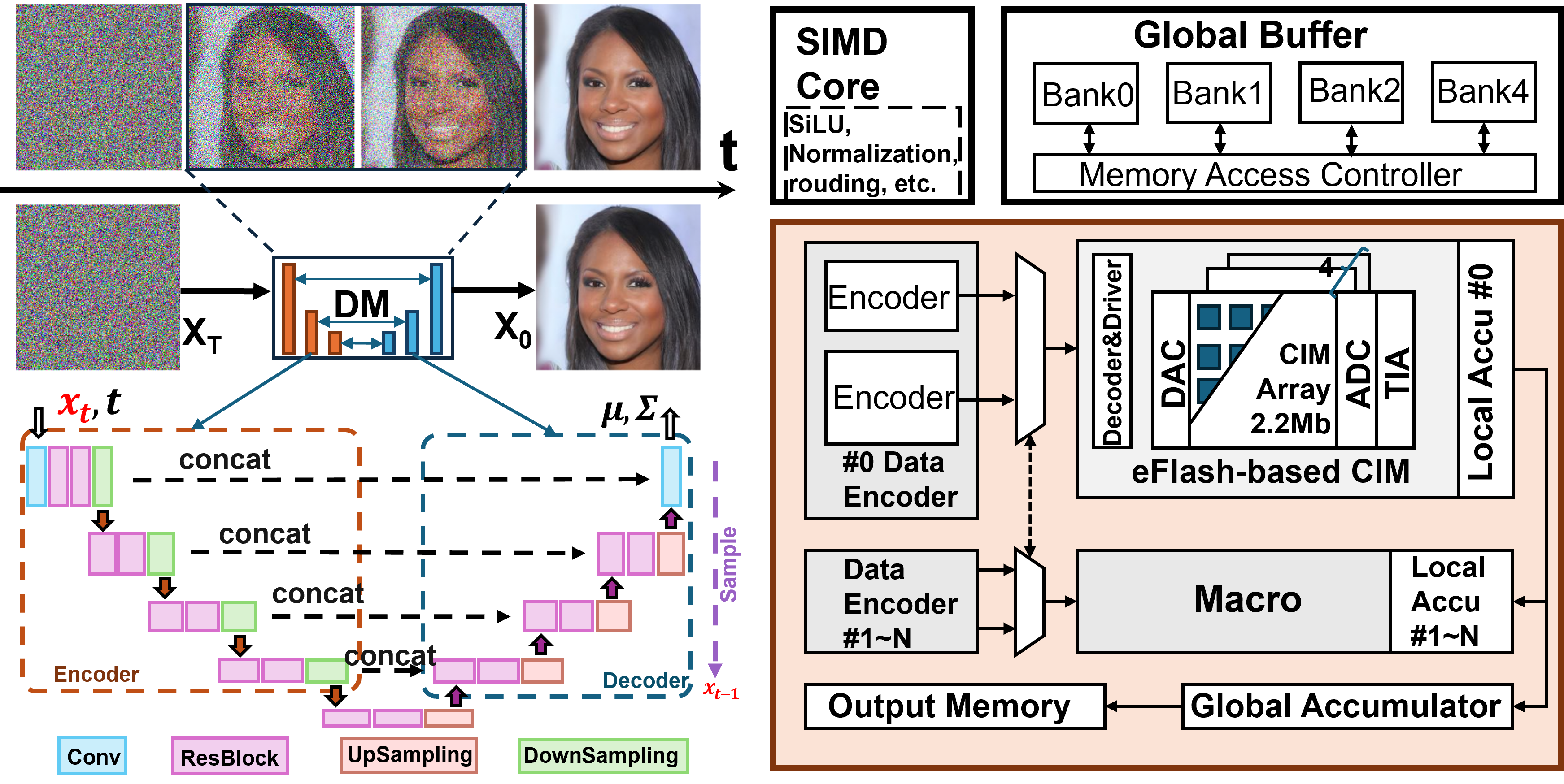}
\caption{Diffusion inference (left) and its CIM acceleration architecture (right).}
% \vspace{-0.2cm}
\label{fig:architecture}
\end{figure}

Despite algorithmic efforts to reduce the computational burden of the iterative denoising process, intensive matrix-vector multiplications (MVMs) remain the primary bottleneck in diffusion model computation. CIM architectures naturally provide energy-efficient solutions for these operations~\cite{watson2021learning,salimans2022progressive,lu2022dpm}, as shown in Figure~\ref{fig:architecture}. Recent hardware advances have demonstrated the feasibility of CIM-based diffusion model acceleration. AIG-CIM~\cite{aigcim} introduces a heterogeneous CIM system to address the flexible data reuse demands in diffusion models, demonstrating more than two orders of magnitude throughput improvement and energy efficiency improvement compared to RTX 3090 GPU. Additionally,~\cite{isscc} presents a heterogeneous CIM chip that supports both integer (INT) and floating-point (FP) arithmetic for accelerating diffusion models while maintaining model performance.

%%%%%%%%%%%%%%%%%%%%%%%%
%%%%%%%%%%%%%%%%%%%%%%%%

\subsection{CIM Noise on Neural Network Robustness}
While analog CIM architectures deliver remarkable gains in energy efficiency and computational throughput, their analog nature inevitably introduces non-idealities and intrinsic noise sources. These factors cause perturbations in stored weights and computed activations, leading to accuracy degradation during inference~\cite{zhou_tcad,bai2024end,asicon1}. Therefore, ensuring model robustness under analog-induced noise has become a crucial challenge for reliable CIM deployment.

Recent works have explored various strategies to improve robustness when deploying large-scale DNNs, including vision transformers (ViTs), Mamba, and Transformer-based LLMs~\cite{joshi2020accurate, liang2021crossbar,reagen2018ares} on CIM hardware. Their methods either require fine-tuning or hardware modifications, which are impractical in the era of exponentially increasing model and dataset sizes. Fine-tuning billion-parameter models demands substantial computational resources and time, while hardware modifications introduce additional design complexity and manufacturing costs. Our training-free approach instead changes only sampler coefficients and random draws: it adds no model parameters, CIM arrays, or network evaluations, although it requires random-number generation and time-dependent sampler control.

%%%%%%%%%%%%%%%%%%%%%%%%%%%%%%%%%%%%%%%%%%%%%%%%%%%%%%%%%%%%%%%%%%%%%%%%%%%%%%%%%%%%%%%%%%%%%%%%%%%%%%%%%%%%%%%%%%
\section{Methodology}
\label{sec:methodology}
% CIM architectures can accelerate diffusion models through in-situ MVMs, but their persistent spatial non-idealities interact with repeated denoising evaluations. We first show that the early high-noise stage is more vulnerable than final refinement, then derive the conditions under which fixed spatial errors accumulate coherently. These observations motivate ASSERT, a training-free schedule that combines early stochastic updates with deterministic refinement.

\subsection{Motivation}
To ensure realistic analysis, our noise modeling is grounded in measurements collected from multiple physical chips, as illustrated in Figure~\ref{fig:chip}. We augment the standard matrix-vector multiplication $y=xW$ with weight perturbations $\epsilon_W \sim \mathcal{N}(0,\sigma_W^2)$ applied to stored weights and computation noise $\epsilon_Y \sim \mathcal{N}(0,\sigma_Y^2)$ applied to the output:
\begin{equation}
y = x(W + \epsilon_W) + \epsilon_Y
\end{equation}
At a fixed operating point, the weight-noise variance depends on weight magnitude:
\begin{equation}
\sigma_W^2 = \alpha_W|W| + b_W
\end{equation}
At each operating point, $(\alpha_W,b_W,\sigma_Y)$ is fitted from several hundred measured chip MVM samples. The fitted weight term aggregates programming variation, leakage, thermal noise, and random telegraph noise; the output term captures ADC error and short-timescale output fluctuation. Each mapped linear or convolutional layer receives an independently sampled weight-perturbation map. This map is fixed throughout one deployment realization and all of its denoising steps, while $\epsilon_Y$ is redrawn for every MVM; repeated runs redraw the deployment realization. Prediction errors across denoising steps therefore share the same dominant spatial perturbation and are generally correlated rather than independently resampled.

Unlike VAEs and GANs, denoising diffusion models repeatedly evaluate the same network. We therefore begin from a temporal perspective. Using the preceding fixed noise realization, we apply the noisy mapping to the linear and convolutional layers of a pre-trained 100-step DDIM sampler over selected denoising intervals and measure the sensitivity of each stage.

\begin{figure}[!t]
\centering
\includegraphics[width=1.0\columnwidth]{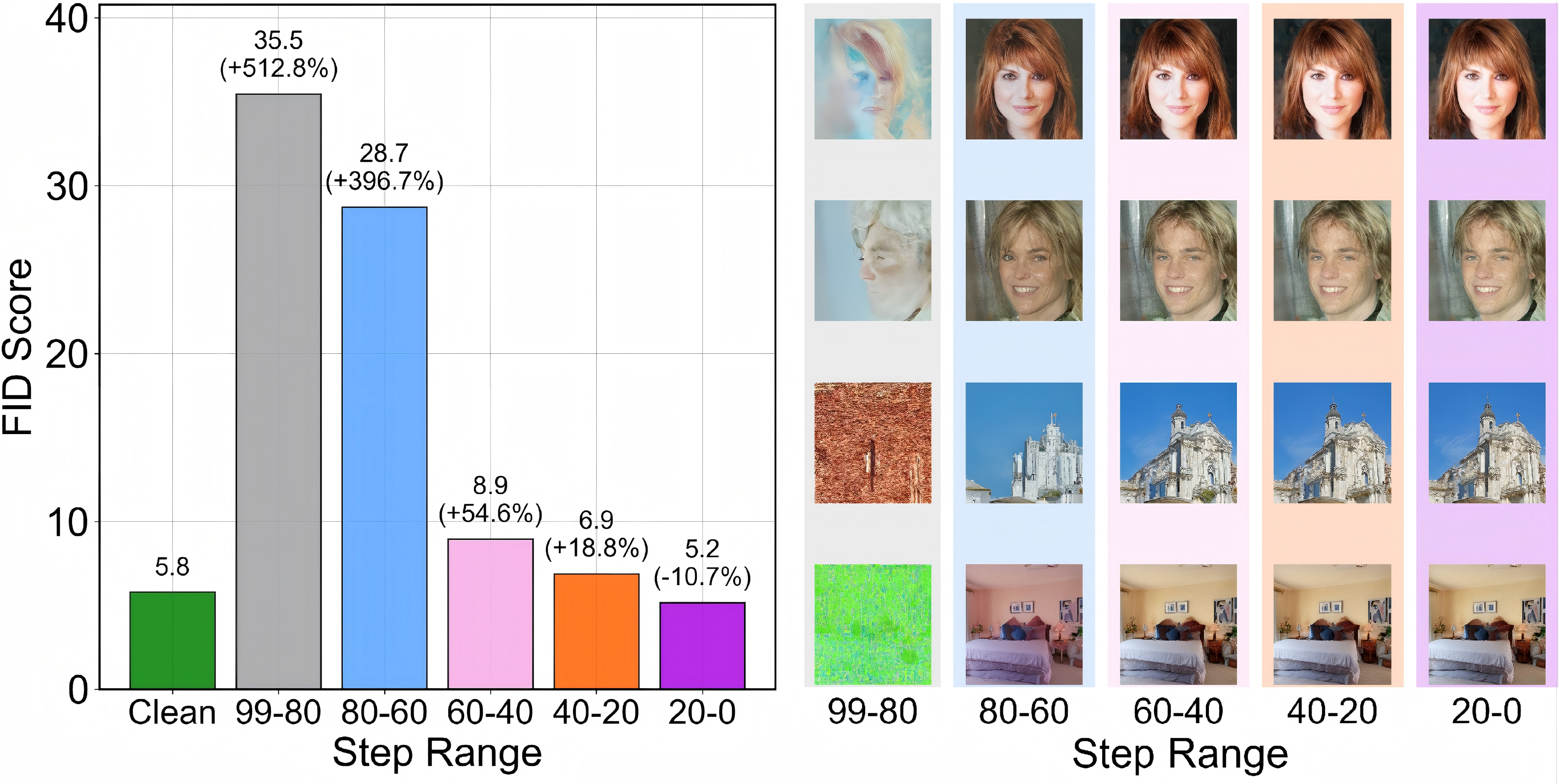}
\caption{Stage-wise FID (left) and samples (right) under fixed hardware noise.}
% \vspace{-0.2cm}
\label{fig:process}
\end{figure}

As shown in Figure~\ref{fig:process}, applying the noisy mapping over the indicated intervals produces markedly different outcomes. Using it during the early high-noise stage (e.g., steps 99--80) gives the largest FID degradation. Applying the same mapping during final refinement (e.g., steps 20--0) has a much smaller effect in this experiment.

The right panel of Figure~\ref{fig:process} presents samples corresponding to the bars in the left panel. This temporal pattern indicates that the early, high-noise stage is particularly important for robustness because perturbations introduced while global structure is formed propagate through all subsequent refinement steps.

DDIM accelerates DDPM sampling by permitting deterministic, timestep-skipping trajectories. We reintroduce controlled stochasticity during the early high-noise stage through $\eta$, where larger values indicate greater stochasticity.

\begin{figure}[!t]
\centering
\includegraphics[width=1.0\columnwidth]{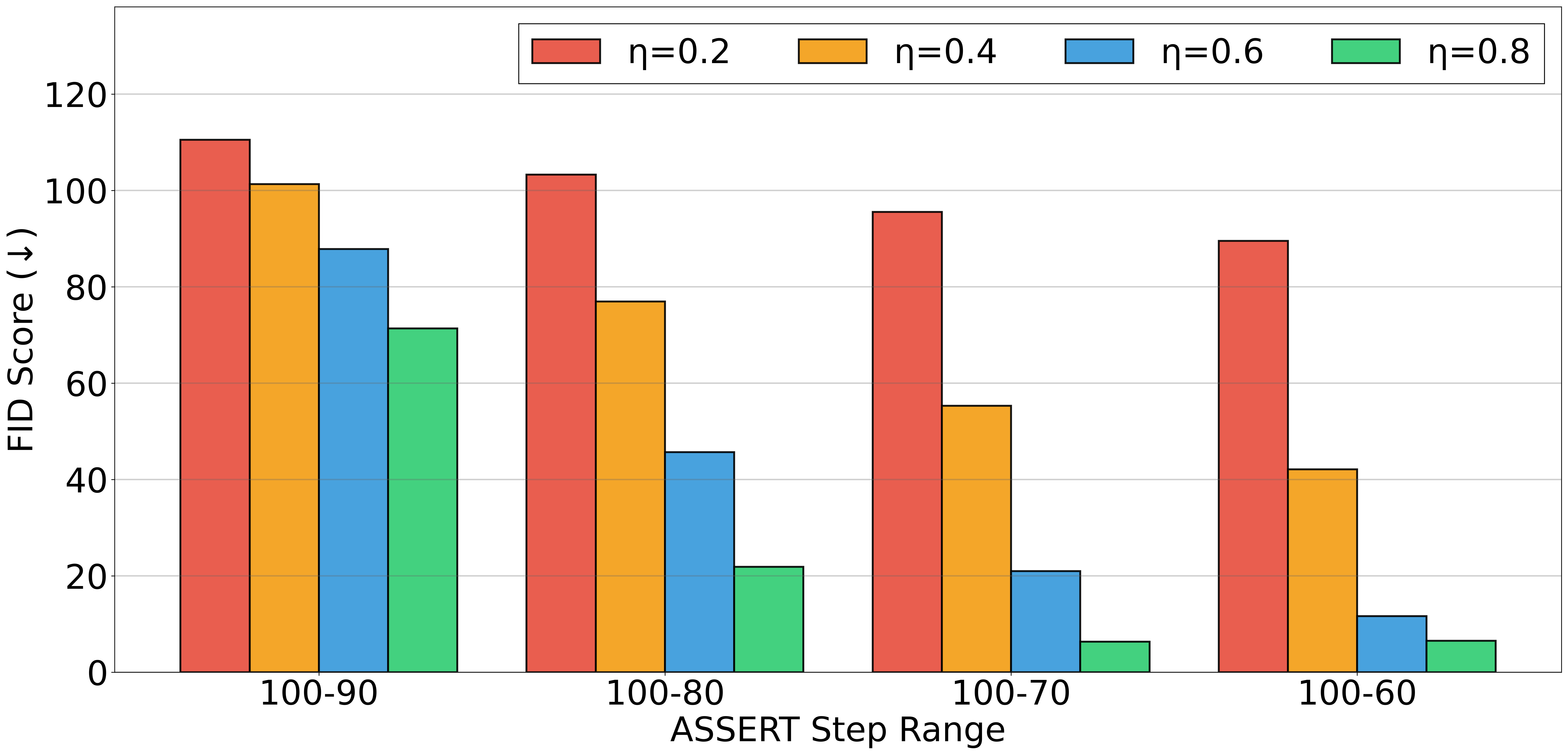}
\caption{FID when constant $\eta$ is active only over each early range.}
% \vspace{-0.2cm}
\label{fig:stepwise}
\end{figure}

Figure~\ref{fig:stepwise} is a diagnostic rather than the complete ASSERT schedule: it applies a constant $\eta$ only inside the indicated early interval and sets $\eta=0$ outside it. Consequently, the limited intervention leaves FID as high as 111 in difficult settings. Increasing $\eta$ and extending the active interval nevertheless reduce FID, with diminishing returns between 100--70 and 100--60 at $\eta=0.8$. Full ASSERT subsequently applies cosine decay throughout the remaining trajectory, producing the much lower FID values in Table~\ref{tab:ablation}. The random update changes the states presented to the same noisy mapping and thereby reduces cross-step error alignment, as formalized in Section~\ref{sec:error_analysis}.
%%%%%%%%%%%%%%%%%%%%%%%%
%%%%%%%%%%%%%%%%%%%%%%%%
%%%%%%%%%%%%%%%%%%%%%%%%
%%%%%%%%%%%%%%%%%%%%%%%%
%%%%%%%%%%%%%%%%%%%%%%%%
\begin{figure*}[]
\centering
\includegraphics[width=1.0\textwidth]{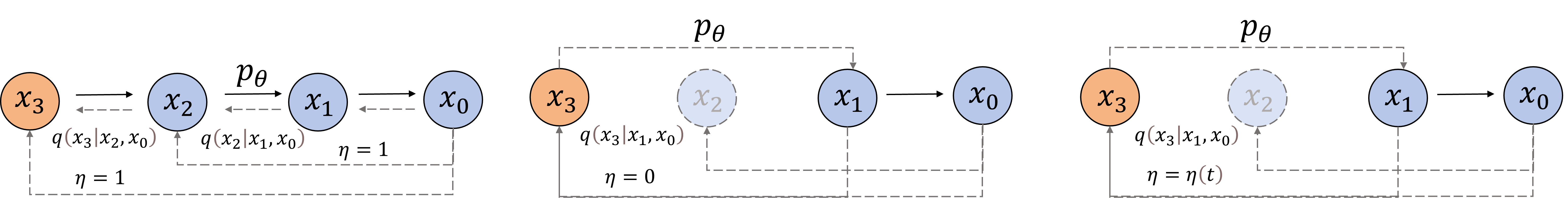}
\caption{Trajectories of stochastic DDIM, deterministic DDIM, and ASSERT.}
% \vspace{-0.2cm}
\label{fig:assert}
\end{figure*}
%%%%%%%%%%%%%%%%%%%%%%%%
\subsection{Persistent-Noise Error Analysis}
\label{sec:error_analysis}
Let $t_0>t_1>\cdots>t_S=0$ be the actual inference subsequence, and consider one transition from $t=t_i$ to $s=t_{i+1}<t$. With $\alpha_t=1-\beta_t$ and $\bar\alpha_t=\prod_{r=1}^{t}\alpha_r$, DDIM predicts
\begin{equation}
\hat{\mathbf{x}}_0=\frac{\mathbf{x}_t-\sqrt{1-\bar\alpha_t}\boldsymbol\epsilon_\theta(\mathbf{x}_t,t)}{\sqrt{\bar\alpha_t}}.
\label{eq:x0_prediction}
\end{equation}
For a dimensionless stochasticity parameter $\eta_i\in[0,1]$, the standard deviation for the selected timestep pair is
\begin{equation}
\sigma_{t\rightarrow s}(\eta_i)=\eta_i
\sqrt{\frac{1-\bar\alpha_s}{1-\bar\alpha_t}}
\sqrt{1-\frac{\bar\alpha_t}{\bar\alpha_s}},
\label{eq:ddim_sigma}
\end{equation}
and the corresponding DDIM update is
\begin{equation}
\mathbf{x}_s=\sqrt{\bar\alpha_s}\hat{\mathbf{x}}_0+
\sqrt{1-\bar\alpha_s-\sigma_{t\rightarrow s}^2}\boldsymbol\epsilon_\theta(\mathbf{x}_t,t)+
\sigma_{t\rightarrow s}\mathbf{z},
\label{eq:general_ddim}
\end{equation}
where $\mathbf{z}\sim\mathcal{N}(0,\mathbf{I})$. Thus, $\eta_i=0$ gives deterministic DDIM, whereas $\eta_i=1$ uses the stochastic DDIM variance associated with the chosen pair $(t,s)$. Importantly, $\eta_i$ is not itself the injected standard deviation.

For fixed $\mathbf{x}_t$ and $\mathbf{z}$, a prediction perturbation $\delta\boldsymbol\epsilon_i$ changes one update by $\delta\mathbf{x}_s=c_i\delta\boldsymbol\epsilon_i$, where
\begin{equation}
c_i=\sqrt{1-\bar\alpha_s-\sigma_{t\rightarrow s}^2}-
\sqrt{\frac{\bar\alpha_s}{\bar\alpha_t}}\sqrt{1-\bar\alpha_t}.
\label{eq:local_coefficient}
\end{equation}
This local coefficient depends on the actual inference subsequence and does not diverge merely because $\bar\alpha_t$ is small. Nevertheless, the clean-sample estimate in Equation~\ref{eq:x0_prediction} has sensitivity
\begin{equation}
\left\|\delta\hat{\mathbf{x}}_0\right\|=
\sqrt{\frac{1-\bar\alpha_t}{\bar\alpha_t}}
\left\|\delta\boldsymbol\epsilon_i\right\|,
\label{eq:x0_sensitivity}
\end{equation}
which increases toward the high-noise end. The cancellation captured by $c_i$ limits the immediate update error, so end-to-end sensitivity must additionally account for how a perturbed state changes all later network evaluations.

Let $f_i$ denote Equation~\ref{eq:general_ddim}, let $\Delta\boldsymbol\theta$ be the fixed spatial weight perturbation of one hardware mapping, and compare noisy and clean trajectories using the same $\mathbf{z}_i$. Let $a_i\in\{0,1\}$ indicate whether the noisy mapping is active at step $i$; the end-to-end evaluation uses $a_i=1$ throughout, whereas Figure~\ref{fig:process} activates it only in one interval. First-order linearization gives
\begin{equation}
\mathbf{e}_{i+1}=\mathbf{J}_i\mathbf{e}_i+a_i\mathbf{H}_i\Delta\boldsymbol\theta,
\quad
\mathbf{J}_i=\frac{\partial f_i}{\partial\mathbf{x}_{t_i}},\quad
\mathbf{H}_i=\frac{\partial f_i}{\partial\boldsymbol\theta}.
\label{eq:error_recursion}
\end{equation}
Unrolling to the final sample yields
\begin{equation}
\mathbf{e}_S=\sum_{i=0}^{S-1}\mathbf{P}_{S,i+1}\mathbf{d}_i,
\quad
\mathbf{d}_i=a_i\mathbf{H}_i\Delta\boldsymbol\theta,\quad
\mathbf{P}_{S,i+1}=\mathbf{J}_{S-1}\cdots\mathbf{J}_{i+1},
\label{eq:unrolled_error}
\end{equation}
with an empty product defined as the identity.
Writing $\mathbf{u}_i=\mathbf{P}_{S,i+1}\mathbf{d}_i$ gives
\begin{equation}
\|\mathbf{e}_S\|^2=\sum_i\|\mathbf{u}_i\|^2+
2\sum_{i<j}\langle\mathbf{u}_i,\mathbf{u}_j\rangle.
\label{eq:correlated_error}
\end{equation}
\textbf{Sufficient condition for early-stage vulnerability.} Because the same $\Delta\boldsymbol\theta$ is reused, all $\mathbf{d}_i$ share one underlying spatial perturbation and their propagated versions can remain directionally aligned. Consider equal-length early and late activation windows with matched local perturbation magnitudes. If (i) the propagation gain $g_i=\|\mathbf{P}_{S,i+1}\mathbf{d}_i\|/\|\mathbf{d}_i\|$ is no smaller for the early window, as occurs when the additional Jacobians are non-contracting on the dominant error subspace, and (ii) $\langle\mathbf{u}_i,\mathbf{u}_j\rangle\geq0$ within each window, then every marginal and cross term in Equation~\ref{eq:correlated_error} is no smaller for the early window. Its terminal error is therefore no smaller. Equation~\ref{eq:x0_sensitivity} further shows why large local perturbations are likely at the high-noise end. These sufficient conditions derive the stage ordering observed in Figure~\ref{fig:process}; they do not require or imply that the single-step coefficient $|c_i|$ diverges. We evaluate the resulting prediction through stage-wise FID, while direct Jacobian-gain and cross-step-cosine measurements are left for future work.

\textbf{Stochastic decorrelation mechanism.} Stochasticity addresses the persistent, directional component rather than directly canceling an additive bias. By changing $\mathbf{x}_{t_i}$, it changes $\mathbf{H}_i$ and hence the direction of $\mathbf{d}_i$ for the same physical $\Delta\boldsymbol\theta$. If stochastic sampling reduces the expected positive cross-step alignment while leaving the marginal terms comparable, Equation~\ref{eq:correlated_error} gives
\begin{equation}
\mathbb{E}\|\mathbf{e}_S\|_{\rm stoch}^2-
\mathbb{E}\|\mathbf{e}_S\|_{\rm det}^2
\approx 2\sum_{i<j}\left(
\mathbb{E}\langle\mathbf{u}_i,\mathbf{u}_j\rangle_{\rm stoch}-
\mathbb{E}\langle\mathbf{u}_i,\mathbf{u}_j\rangle_{\rm det}\right)<0.
\label{eq:decorrelation}
\end{equation}
This explains why stochastic updates can mitigate fixed spatial noise even though independent randomness cannot remove a fixed bias in a single step. Excess stochasticity can still increase marginal error, motivating a schedule that concentrates it in the vulnerable early stage. The stage and schedule experiments test this mechanism at the output-distribution level; direct internal-error correlation is not measured in the current study.

\subsection{ASSERT Framework}
ASSERT applies Equation~\ref{eq:general_ddim} without retraining or additional network evaluations. As illustrated in Figure~\ref{fig:assert}, the dimensionless $\eta_i$ controls the transition from stochastic exploration to deterministic refinement, while Equation~\ref{eq:ddim_sigma} converts it to a valid pair-dependent standard deviation.

For $S>1$ sampling steps, define normalized denoising progress $r_i=i/(S-1)$, where $i=0$ is the early high-noise step. ASSERT uses
\begin{equation}
\eta_i = \begin{cases}
\eta_{\max}, & r_i < \tau_{\text{trans}}, \\
\eta_{\max}\mathcal{D}\left(\frac{r_i-\tau_{\text{trans}}}{1-\tau_{\text{trans}}}\right), & r_i \geq \tau_{\text{trans}},
\end{cases}
\label{eq:assert_schedule}
\end{equation}
where $\tau_{\text{trans}}\in[0,1)$ is the fraction of steps held at maximum stochasticity and
\begin{equation}
\mathcal{D}(p)=\frac{1+\cos(\pi p)}{2}.
\end{equation}
The use of $S-1$ ensures that the last update reaches $\eta_{S-1}=0$. The two hyperparameters $\eta_{\max}$ and $\tau_{\text{trans}}$ therefore specify the stochasticity magnitude and its early plateau, respectively.
%%%%%%%%%%%%%%%%%%%%%%%%%%%%%%%%%%%%%%%%%%%%%%%

\section{Experiments}
\label{sec:experiments}
\subsection{Experiment Setup}
%%%%%%%%%%%%%%%%%%%%%%%%
\begin{figure}[htbp]
\centering
\includegraphics[width=1.0\columnwidth]{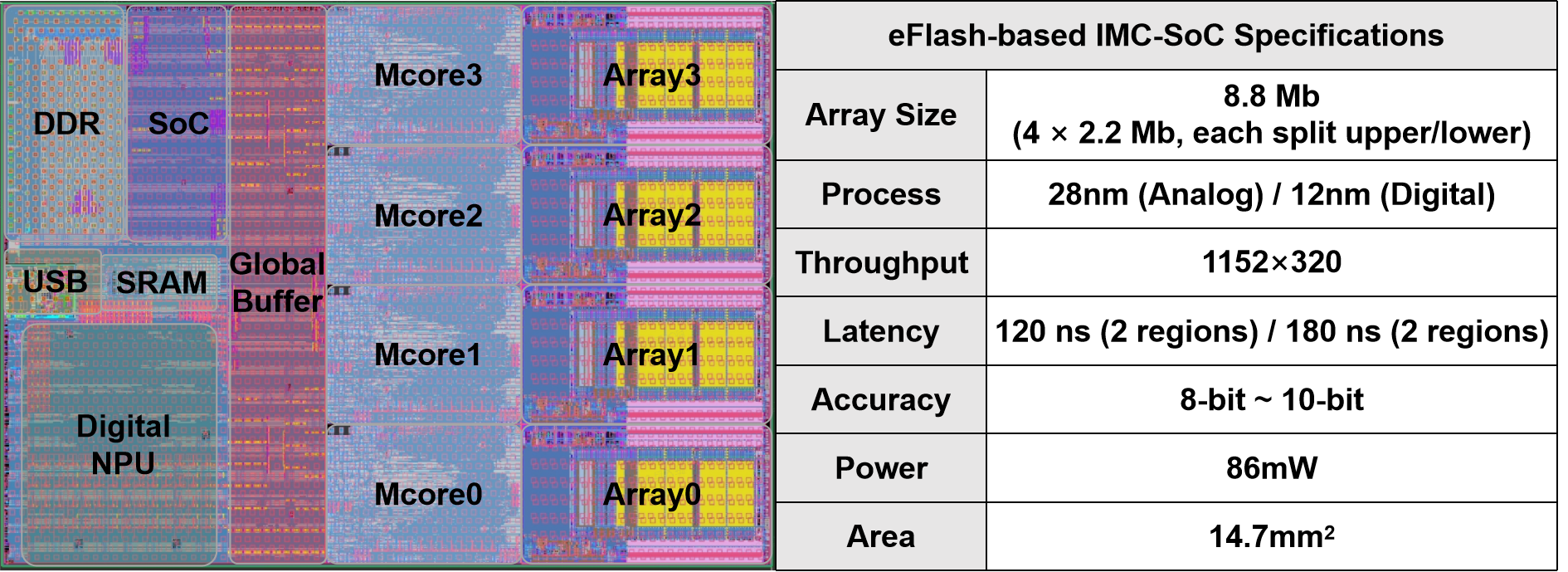}
\caption{Measured eFlash CIM chip and SoC specifications.}
% \vspace{-0.2cm}
\label{fig:chip}
\end{figure}
%%%%%%%%%%%%%%%%%%%%%%%% 
We evaluated public Hugging Face diffusion checkpoints for CIFAR-10 (32$\times$32), Butterfly (64$\times$64), CelebA-HQ (256$\times$256), LSUN-Church (256$\times$256), and LSUN-Bedroom (256$\times$256). All experiments use the released exponential-moving-average weights, BF16 inference, and each checkpoint's native prediction parameterization and training noise schedule.

Following DDIM~\cite{DDIM}, an $S$-step run uniformly subsamples a training horizon of $T$ steps as $\{\lfloor jT/S\rfloor\}_{j=0}^{S-1}$ and traverses it in descending order. We select $\eta_{\max}=0.8$ and a 10\% early plateau from the CIFAR-10 ablation, then keep both values fixed across datasets. The dimensionless schedule is shared, while $\sigma_{t\rightarrow s}(\eta_i)$ is recomputed for every selected timestep pair.

Our noise model was fitted to measurements collected across physical eFlash CIM chips and validated on measurements from 100 chips. Figure~\ref{fig:chip} shows the shared 28-nm platform, which integrates four 2.2-Mb CIM arrays with 8.8 Mb total capacity, 86 mW power, and 14.7 mm$^2$ area. Linear and convolutional MVMs use independently sampled analog weight maps, with larger tensors tiled across arrays; normalization, nonlinear activation, residual addition, and sampler control remain digital. We evaluate the complete diffusion trajectory end to end with this mapping, rather than claiming that the entire model executes on one physical chip.

Each FID value uses 50,000 generated images and the standard reference statistics associated with the corresponding public dataset. We report the mean over three seeds. Within each seed, all compared samplers share the same initial latents and fixed deployment-noise realization; different seeds redraw both. The reported hardware noise level is the equivalent weight-noise standard deviation, distinct from the sampler parameter $\eta$ and from a direct temperature label.

\subsection{Experiment Results}
To comprehensively evaluate our adaptive sampling strategy, we compare ASSERT against several baseline sampling approaches with different stochasticity scheduling patterns. Linear decay gradually reduces stochasticity from maximum to minimum following 
\begin{equation}
\eta_i = \eta_{\max} - \frac{i}{S-1} \cdot (\eta_{\max} - \eta_{\min})
\end{equation}
which simplifies to $\eta_i = \eta_{\max} \left(1-i/(S-1)\right)$ when $\eta_{\min}=0$. Step decay (left) and reverse step (right) represent two opposite scheduling patterns:
\begin{equation}
\eta_i = \begin{cases} 
\eta_{\max}, & \text{if } r_i < \tau_{\text{trans}} \\
\eta_{\min}, & \text{if } r_i \geq \tau_{\text{trans}}
\end{cases}  \quad
\eta_i = \begin{cases} 
\eta_{\min}, & \text{if } r_i < \tau_{\text{trans}} \\
\eta_{\max}, & \text{if } r_i \geq \tau_{\text{trans}}
\end{cases}
\end{equation}
where $r_i=i/(S-1)$ and the evaluated setting uses $\tau_{\text{trans}}=0.4$. These schedules isolate whether stochasticity is most useful during early denoising or final refinement.

\begin{figure}[!t]
\centering
\includegraphics[width=1.0\columnwidth]{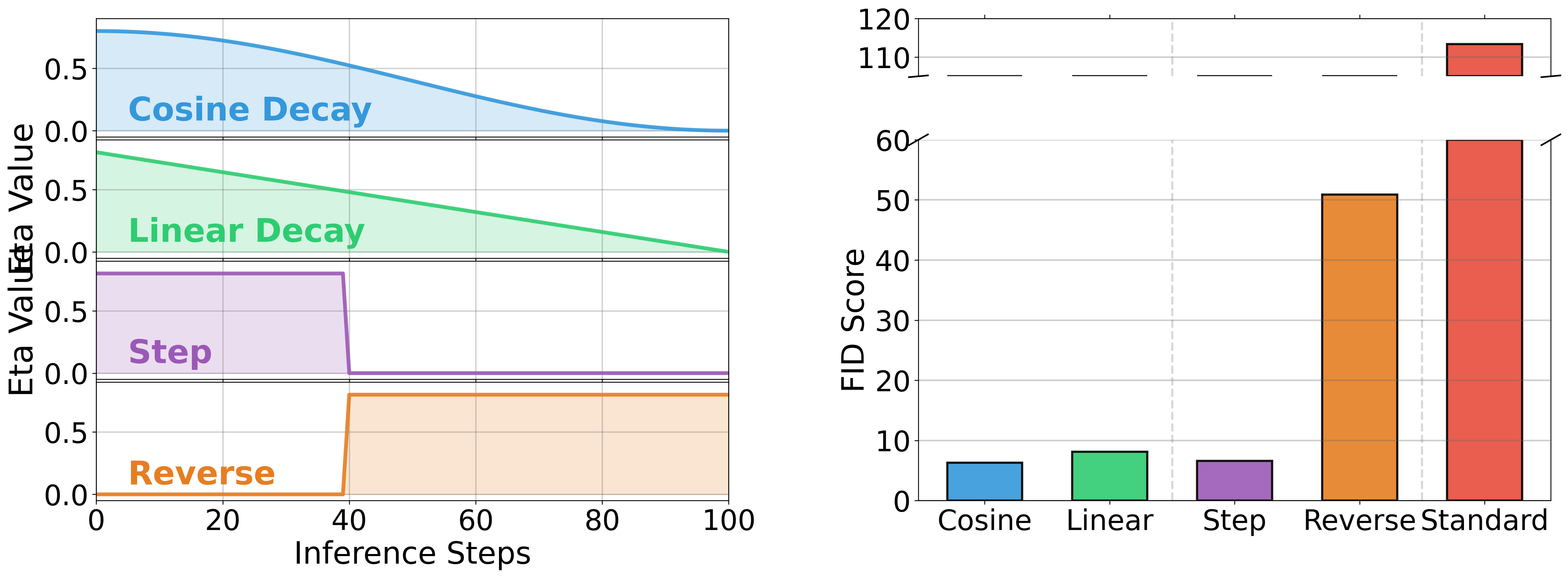}
\caption{Stochasticity schedules (left) and noisy CIFAR-10 FID (right).}
% \vspace{-0.2cm}
\label{fig:diffdecay}
\end{figure}

\textbf{Decay Strategies Comparison.} The left panel of Figure~\ref{fig:diffdecay} visualizes the schedules, and the right panel reports their CIFAR-10 FID under the evaluated hardware-noise setting. Cosine decay obtains the lowest FID, followed by step and linear decay. Reverse step, which reserves stochasticity for final refinement, performs much worse than the early-stochastic schedules. Its small gain over deterministic DDIM indicates that stochasticity alone can help, while the large gap to ASSERT supports concentrating it in the vulnerable early stage.

\begin{figure}[!t]
\centering
\includegraphics[width=1.0\columnwidth]{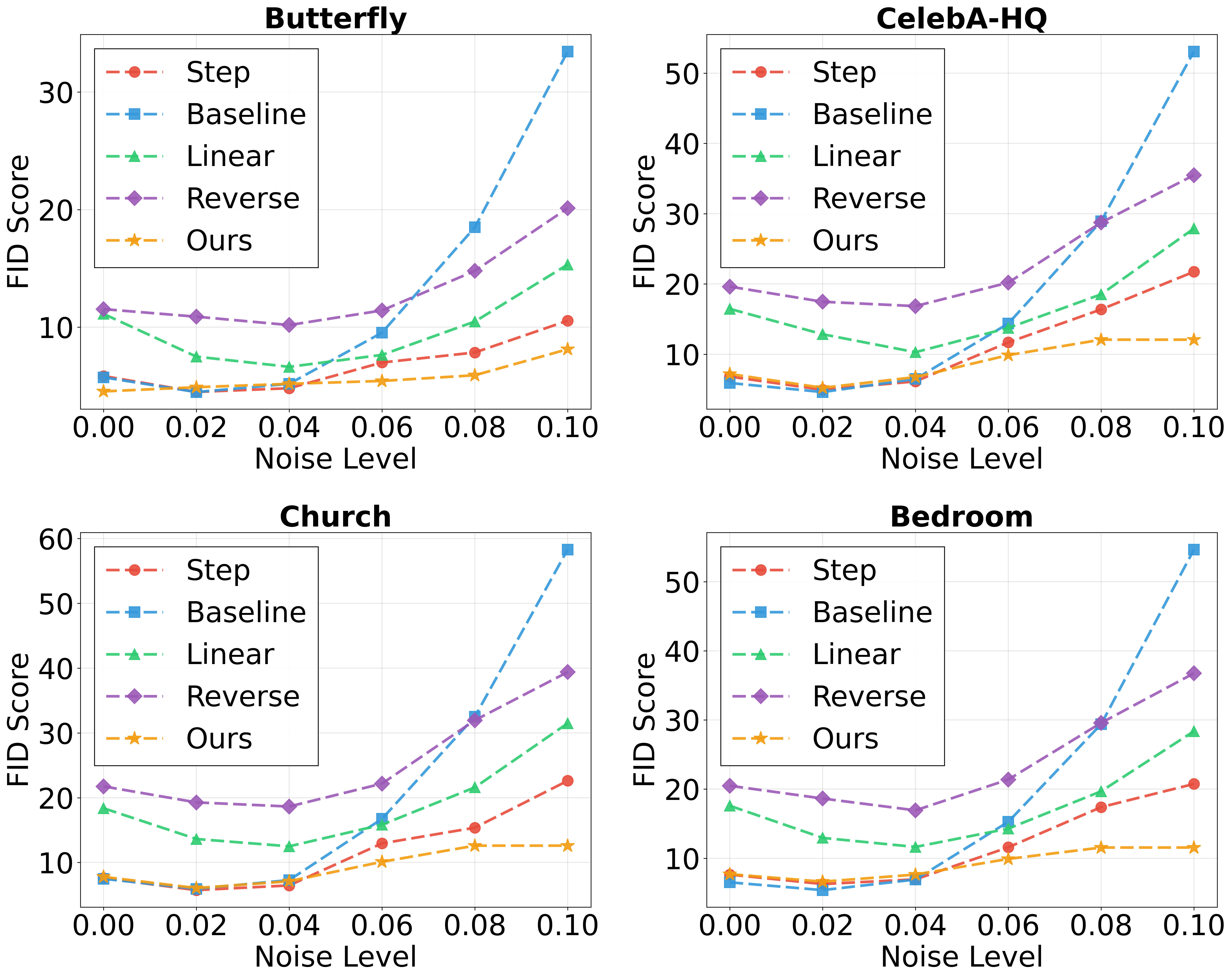}
\caption{FID across datasets, samplers, and weight-noise levels.}
% \vspace{-0.2cm}
\label{fig:fid_comparison_4datasets}
\end{figure}

\textbf{Evaluation Across Datasets.} Figure~\ref{fig:fid_comparison_4datasets} extends the comparison to four datasets and several equivalent weight-noise levels. At noise level 0.08, ASSERT obtains $2.39\times$, $2.58\times$, and $2.54\times$ lower mean FID than deterministic DDIM on CelebA-HQ, Church, and Bedroom, respectively. Relative to the step schedule, the corresponding reduction factors are $1.36\times$, $1.22\times$, and $1.51\times$. These ratios use the three-seed mean FID and are not absolute FID differences.
%%%%%%%%%%%%%%%%%%%%%%%%
%%%%%%%%%%%%%%%%%%%%%%%%
%%%%%%%%%%%%%%%%%%%%%%%%
%%%%%%%%%%%%%%%%%%%%%%%%
%%%%%%%%%%%%%%%%%%%%%%%%
\begin{figure}[!t]
\centering
\includegraphics[width=1.0\columnwidth]{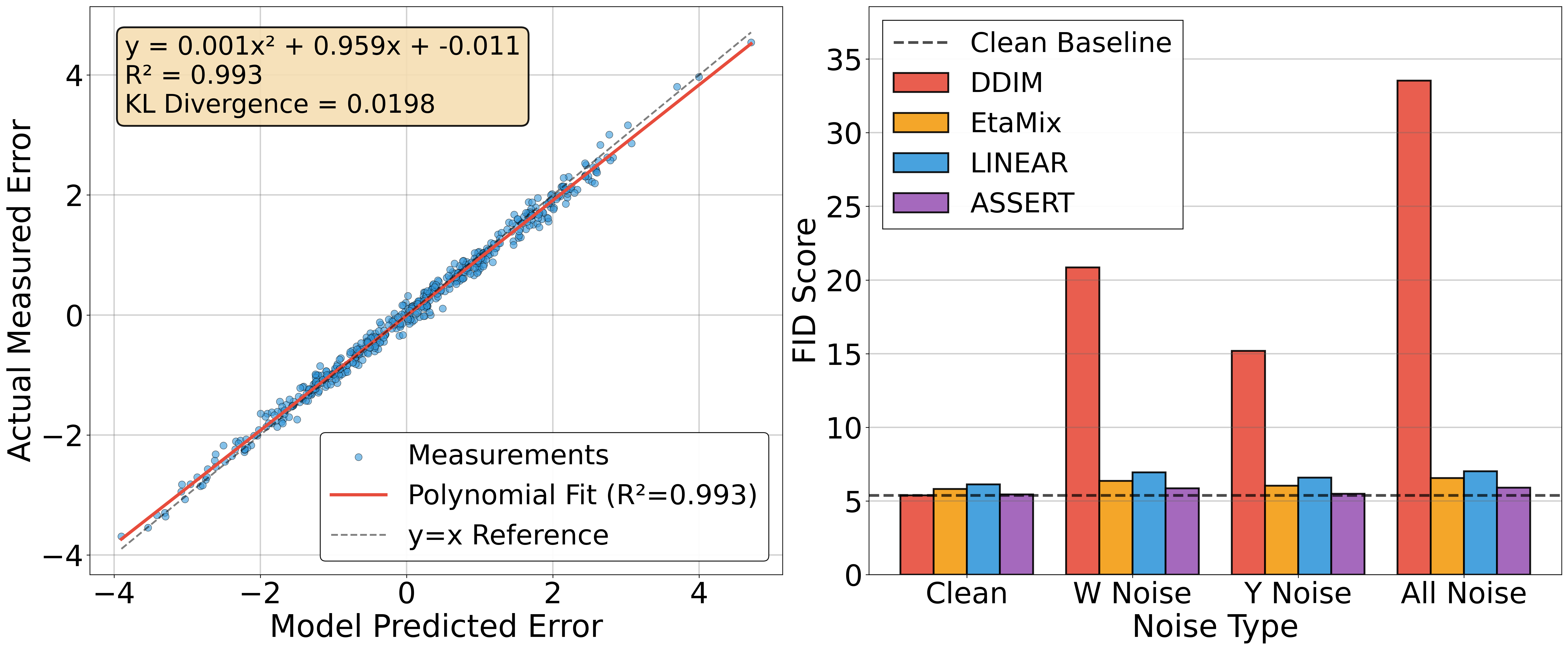}
\caption{Chip-error validation (left) and FID by noise source (right).}
% \vspace{-0.2cm}
\label{fig:diffnoise_combined_fid_comparison}
\end{figure}
%%%%%%%%%%%%%%%%%%%%%%%%

\textbf{Noise Model Validation.} The left panel of Figure~\ref{fig:diffnoise_combined_fid_comparison} compares model-predicted errors with measurements aggregated from 100 physical chips, obtaining $R^2=0.993$. We compute $D_{\mathrm{KL}}(p_{\mathrm{chip}}\|p_{\mathrm{model}})=0.0198$ from normalized measured and modeled error histograms with shared bin edges. The displayed polynomial is a goodness-of-fit diagnostic, not an alternative to the variance model above. The right panel applies weight noise, output noise, and their combination throughout the complete diffusion trajectory. ASSERT remains close to the clean baseline, whereas deterministic DDIM degrades most strongly under persistent weight noise, consistent with fixed spatial variations creating correlated errors across denoising steps.
%%%%%%%%%%%%%%%%%%%%%%%%
\begin{table}[!t]
\renewcommand{\arraystretch}{1.25}
\caption{CIFAR-10 FID across denoising step counts.}
\centering
\begin{tabular}{lcccccc}
\toprule
\multirow{2}{*}{Steps} & \multicolumn{2}{c}{DDIM} & \multicolumn{2}{c}{STEP} & \multicolumn{2}{c}{ASSERT} \\ 
\cmidrule(lr){2-3} \cmidrule(lr){4-5} \cmidrule(lr){6-7}
 & Clean & Noise & Clean & Noise & Clean & Noise \\ 
\midrule
10 & 18.68 & 29.37 & 20.23 & 24.52 & 19.88 & \cellcolor{blue!20}20.16 \\
20 & 11.08 & 17.19 & 11.54 & 14.41 & 13.98 & \cellcolor{blue!20}13.69 \\
50 & 7.18 & 16.28 & 7.35 & 9.60 & 7.39 & \cellcolor{blue!20}7.44 \\
100 & 5.39 & 33.53 & 5.81 & 6.55 & 5.45 & \cellcolor{blue!20}5.90 \\
200 & 4.84 & 89.66 & 4.96 & 19.45 & 4.90 & \cellcolor{blue!20}15.29 \\
500 & 4.31 & 163.38 & 4.41 & 23.88 & 4.45 &  \cellcolor{blue!20}21.27 \\
1000 & 4.12 & 182.61 & 4.15 & 96.11 & 4.20 & \cellcolor{blue!20}80.39 \\
\bottomrule
\end{tabular}

\label{tab:fid_steps}
\end{table}
%%%%%%%%%%%%%%%%%%%%%%%%
%%%%%%%%%%%%%%%%%%%%%%%%
\begin{figure}[!t]
\centering
\includegraphics[width=1.0\columnwidth]{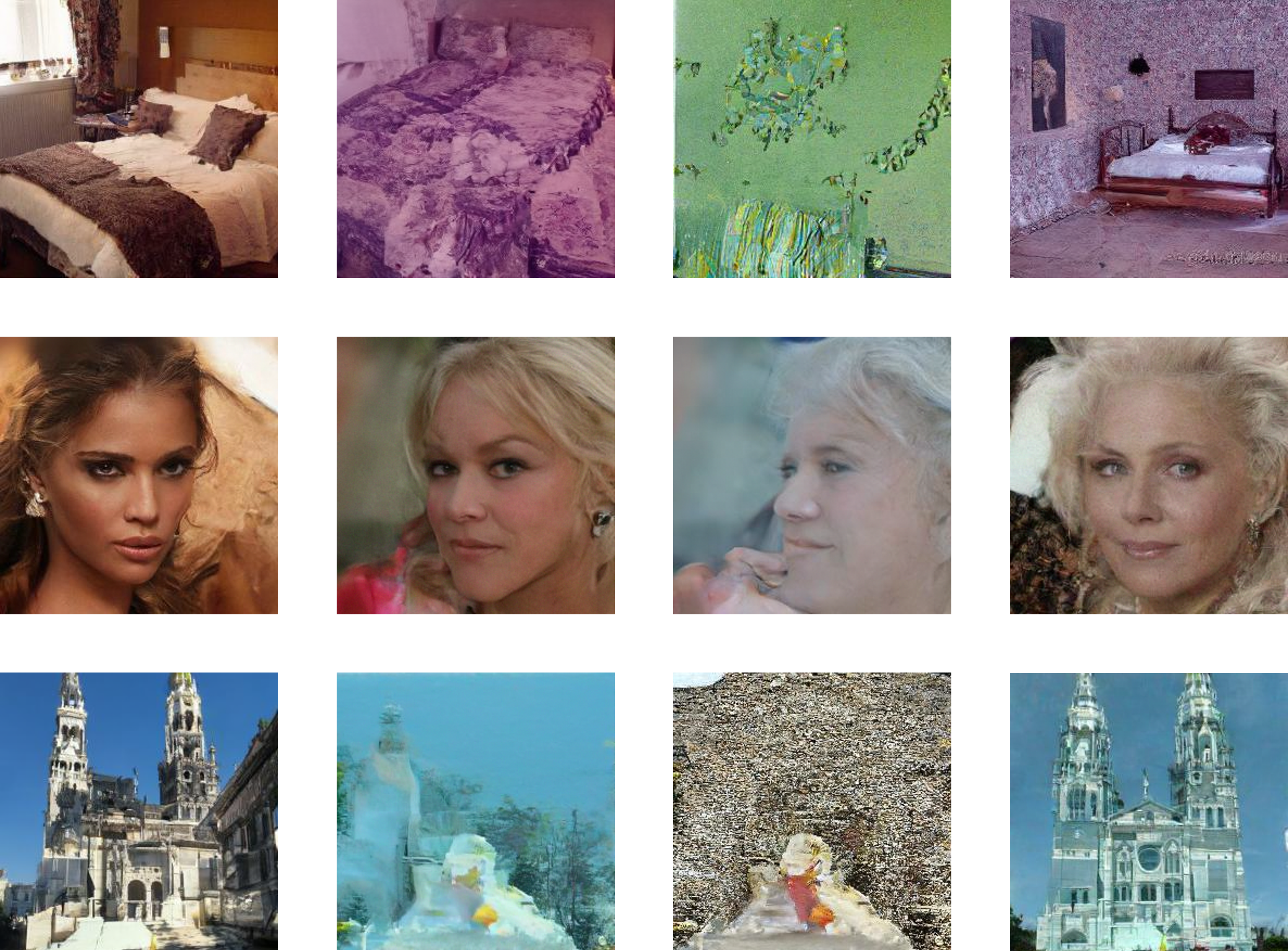}
\caption{Bedroom, CelebA-HQ, and Church samples under clean and noisy inference.}
% \vspace{-0.2cm}
\label{fig:visualization}
\end{figure}
%%%%%%%%%%%%%%%%%%%%%%%%

\textbf{Sampling Strategy Comparison.} Table~\ref{tab:fid_steps} compares deterministic DDIM, STEP, and ASSERT across different step counts. ASSERT has the lowest noisy FID in every row. The largest ratio occurs at 500 steps, where FID decreases from 163.38 to 21.27, corresponding to $163.38/21.27=7.68\times$ lower FID. At 100 steps, it decreases from 33.53 to 5.90 ($5.68\times$). This robustness is not free of clean-quality tradeoffs: for example, 20-step ASSERT has clean FID 13.98 versus 11.08 for DDIM. The noisy result is also non-monotonic in step count, indicating that repeated exposure to a fixed hardware perturbation competes with the benefit of additional denoising. The 20-step noisy ASSERT value (13.69) being slightly below its clean value requires repeated-seed validation before it can be interpreted as an improvement.

\textbf{Visual Comparison.} Figure~\ref{fig:visualization} compares Bedroom, CelebA-HQ, and Church samples under the same four sampling settings. In these examples, noisy DDIM exhibits color distortion and detail loss, whereas ASSERT better preserves facial structure and scene texture. This qualitative comparison complements, but does not replace, the aggregate FID evaluation.
%%%%%%%%%%%%%%%%%%%%%%%%
\begin{table}[!t]
\renewcommand{\arraystretch}{1.5}
\caption{CIFAR-10 FID versus $\eta_{\max}$ and cosine-decay onset.} 
\centering
\begin{tabular}{lccccc}
\toprule
\multirow{2}{*}{Onset} & \multicolumn{5}{c}{FID ($\downarrow$) by $\eta_{\max}$} \\ \cmidrule(lr){2-6}
 & $0.6$ & $0.7$ & $0.8$ & $0.9$ & $1.0$ \\ \midrule
100 & 17.08 & 9.20 & 6.27 & 6.48 & 7.77 \\
90 & 11.48 & 6.28 &  \cellcolor{red!20}5.95 & 6.69 & 7.58 \\
80 & 10.55 & 6.57 & 6.53 & 7.67 & 8.88 \\
70 & 9.53 & 6.47 & 6.65 & 7.70 & 8.81 \\
60 & 9.28 & 6.41 & 6.54 & 8.31 & 8.52 \\
\bottomrule
\end{tabular}
\label{tab:ablation}
% \vspace{-0.2cm}
\end{table}
%%%%%%%%%%%%%%%%%%%%%%%%%%%%%%%%%%%%%%%%%%%%%
\subsection{Ablation Study}
Table~\ref{tab:ablation} studies the maximum stochasticity and the denoising timestep at which cosine decay begins. Because sampling proceeds from 100 toward 0, onset 90 means that $\eta_{\max}$ is held for the first 10 steps, whereas onset 60 holds it for the first 40 steps; onset 100 starts decay immediately. For this 100-step setting, the normalized parameter in Equation~\ref{eq:assert_schedule} is $\tau_{\text{trans}}=(100-\text{onset})/99$. With $\eta_{\max}=0.6$, a longer plateau steadily lowers FID from 17.08 to 9.28. At larger $\eta_{\max}$, excessive exposure is counterproductive: the best observed value is 5.95 at $\eta_{\max}=0.8$ and onset 90. This interaction supports a short, strong stochastic phase followed by gradual deterministic refinement rather than maximizing either hyperparameter independently.

%%%%%%%%%%%%%%%%%%%%%%%%%%%%%%%%%%%%%%%%%%%%%%%%%%%%%%%%%%%%%%%%%%%%%

\section{Conclusion}
We investigated diffusion inference under persistent spatial noise using a model calibrated and validated against measurements from multiple physical CIM chips. A first-order trajectory recursion shows that fixed hardware perturbations create correlated prediction errors whose early contributions propagate through more subsequent updates; stochastic updates can reduce the positive cross-step alignment by changing the activation trajectory. Based on this mechanism, ASSERT applies valid pair-dependent stochastic DDIM updates early and smoothly decays to deterministic refinement without retraining or additional network evaluations. ASSERT achieves up to 2.58$\times$ lower FID than deterministic DDIM on the evaluated high-resolution datasets and 7.68$\times$ lower FID in the CIFAR-10 step-count study. These results support adaptive stochasticity as an algorithmic robustness mechanism for diffusion inference with measured CIM non-idealities.

\bibliographystyle{acm}
\bibliography{bibliography}
\end{document}